\documentclass[10pt,twocolumn,letterpaper]{article}

\usepackage{cvpr}          

\usepackage{graphicx}
\usepackage{amsmath}
\usepackage{amssymb}
\usepackage{booktabs}
\usepackage{times}
\usepackage{float}
\usepackage{graphicx}
\usepackage{amsmath}
\usepackage{amssymb}
\usepackage{array}
\usepackage[toc,page]{appendix}
\usepackage{subcaption}
\usepackage{multirow}
\usepackage{tabularx}
\usepackage{makecell}
\usepackage{adjustbox}
\usepackage{url}
\usepackage[dvipsnames]{xcolor}
\usepackage[accsupp]{axessibility}
\usepackage{xcolor,color,framed}
\usepackage{url, overpic, cases, contour}
\usepackage[symbol]{footmisc}
\usepackage[pagebackref,breaklinks,colorlinks]{hyperref}
\usepackage[capitalize]{cleveref}
\crefname{section}{Sec.}{Secs.}
\Crefname{section}{Section}{Sections}
\Crefname{table}{Table}{Tables}
\crefname{table}{Tab.}{Tabs.}

\def\confName{CVPR}
\def\confYear{2022}

\begin{document}

\title{Neural Compression-Based Feature Learning for Video Restoration}

\author{Cong Huang$^1\thanks{ This work was done when Cong Huang was an intern at Microsoft Research Asia. }\qquad\!$ Jiahao Li $^2\qquad\!$ Bin Li$^2 \qquad\!$ Dong Liu$^1 \qquad\!$ Yan Lu$^2$\\
$^1$ University of Science and Technology of China $\ \ ^2$ Microsoft Research Asia\\
{\tt\small hcy96@mail.ustc.edu.cn, dongeliu@ustc.edu.cn, \{li.jiahao, libin, yanlu\}@microsoft.com}
}
\maketitle

\begin{abstract}
How to efficiently utilize the temporal features is crucial, yet challenging, for video restoration. The temporal features usually contain various noisy and uncorrelated  information, and they may interfere with the restoration of the current frame. This paper proposes learning noise-robust feature representations to help video restoration. We are inspired by that the neural codec is a natural denoiser. In neural codec, the  noisy and uncorrelated  contents which are hard to predict but cost lots of bits are more inclined to be discarded for bitrate saving. Therefore,   we design a neural compression module to filter the noise  and keep the most useful information in features for video restoration. To achieve robustness to noise, our compression module adopts a spatial-channel-wise quantization mechanism to adaptively determine the quantization step size for each position in the latent. Experiments show that our method can significantly boost the performance on video denoising, where we obtain 0.13 dB improvement over BasicVSR++ with only 0.23x FLOPs. Meanwhile, our method also obtains SOTA results on video deraining and dehazing.

\end{abstract}

\section{Introduction}

\begin{figure}[t]
		\begin{center}
			\includegraphics[width=1.0\linewidth]{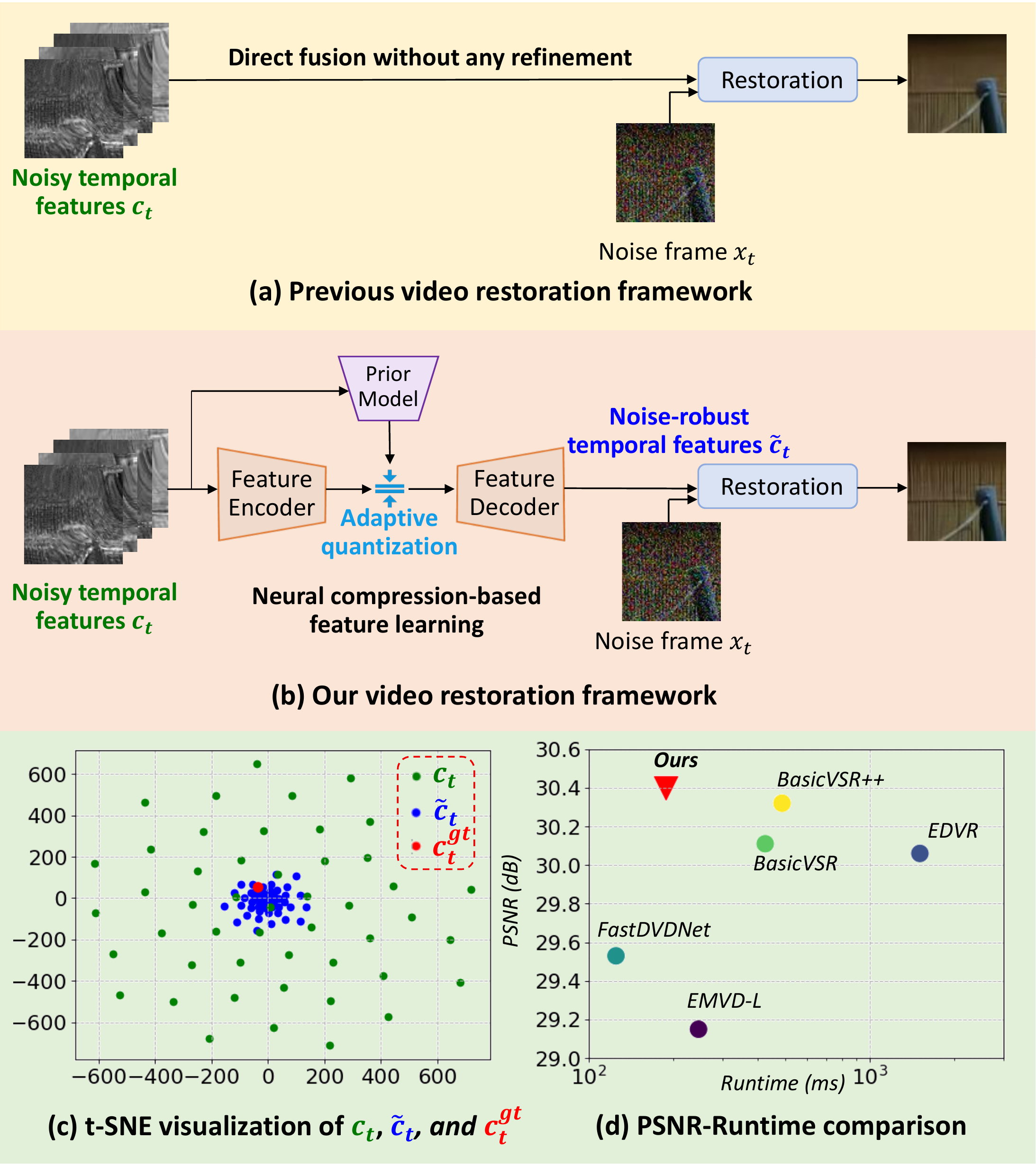}
		\end{center}
		\caption{ 
		(a) Previous framework without temporal feature refinement.
		(b) Our framework via  neural compression-based feature learning.
		(c) t-SNE \cite{tsne} visualization. $c^{gt}_t$ are sampled from clean video (\textit{hypersmooth}, Set8~\cite{fastdvd}).
		For $c_t$ and $\tilde{c}_{t}$, we add different additive white Gaussian noises (same noise $\sigma$ but different noise random seeds) to the same input video to sample these feature points. It shows that $\tilde{c}_{t}$ are more robust to noise and get closer to $c^{gt}_t$.
		(d) Performance comparison on video denoising (Set8, noise $\sigma = 50$). }
		\vspace{-3mm}
		\label{runtime}
	\end{figure}

 Video restoration aims to recover the high-quality video from the degraded input. Typical degradation includes various noises, rain, haze, etc. It has a wide range of applications,  but this problem is still under-explored. 
Different from image restoration that focuses on the intrinsic proprieties in single image~\cite{mprnet}, video restoration relies more on extracting and utilizing temporal features for better quality.

Recent video restoration methods mainly focus on network structure design for better extracting temporal features. For example, RViDeNet~\cite{crvd} and EDVR~\cite{edvr} use deformable convolution to align the features of neighboring frames.
BasicVSR \cite{basicvsr} designs a bi-directional feature propagation network. 
BasicVSR++ \cite{basicvsrpp} introduces the second-order grid propagation network structure and flow-guided deformable alignment network. However, these methods directly use the extracted temporal features without any refinement. The temporal features usually contain lots of noisy and irrelevant information, which interferes with the restoration of the current frame. In this paper, we take video denoising as a case study and explore how to   utilize the extracted temporal features efficiently.

We propose a novel neural compression-based solution to refine the features and learn  noise-robust feature representations. 
From the perspective of neural codec, the noisy data usually contains lots of high-frequency and is hard to predict. To save the bitrate, codec prefers to discard these noisy and uncorrelated  contents. 
This motivates us to design a neural compression module to purify the temporal features and filter the noisy information therein for video restoration. 
To achieve robustness to noise, namely let the representations of the noise-perturbed data be mapped to the same quantized representation with the clean data with high probability, the quantization step needs to be properly set. However, most existing neural compression frameworks only support fixed quantization step size. This cannot meet our purpose and even  harms the inherent textures. To solve this problem,  we design an adaptive quantization mechanism at spatial-channel-wise for our compression module, where the quantization step is learned by our prior model.  Our quantization mechanism can adaptively purify the features with different content characteristics.
During the training, the cross-entropy loss is used to guide the learning of the compression module and helps preserve the most useful information.

Fig.~\ref{runtime} shows the framework comparison. From the t-SNE~\cite{tsne} visualization shown in Fig.~\ref{runtime} (c), we find, via our neural compression-based feature learning, the features are more robust to noise and get closer to the features generated from the clean video. Fig.~\ref{runtime} (d) is the performance comparison. We observe that, empowered by the noise-robust feature representations, our framework  significantly improves the restoration quality, when compared with prior state-of-the-art (SOTA) methods. 
The major contributions of this paper are summarized as follows:

$\bullet$ We propose a novel neural compression-based feature learning for video restoration. After processed by our compression module, the features are more robust to noise and then improve the restoration quality. 

$\bullet$ To achieve robustness to noise and adaptively purify the features with different content characteristics, we design a learnable quantization mechanism at spatial-channel-wise. 

$\bullet$ To further boost the performance, we also design an attention module to  help the feature learning, and a motion vector refinement module to improve the discontinuous motion vector estimated from noisy video.

$\bullet$ We propose a lightweight framework. Compared with previous SOTA methods, our method achieves a better quality-complexity trade-off on video denoising, deraining, and dehazing.

\begin{figure*}
		 \begin{center}
		\includegraphics[width=\linewidth]{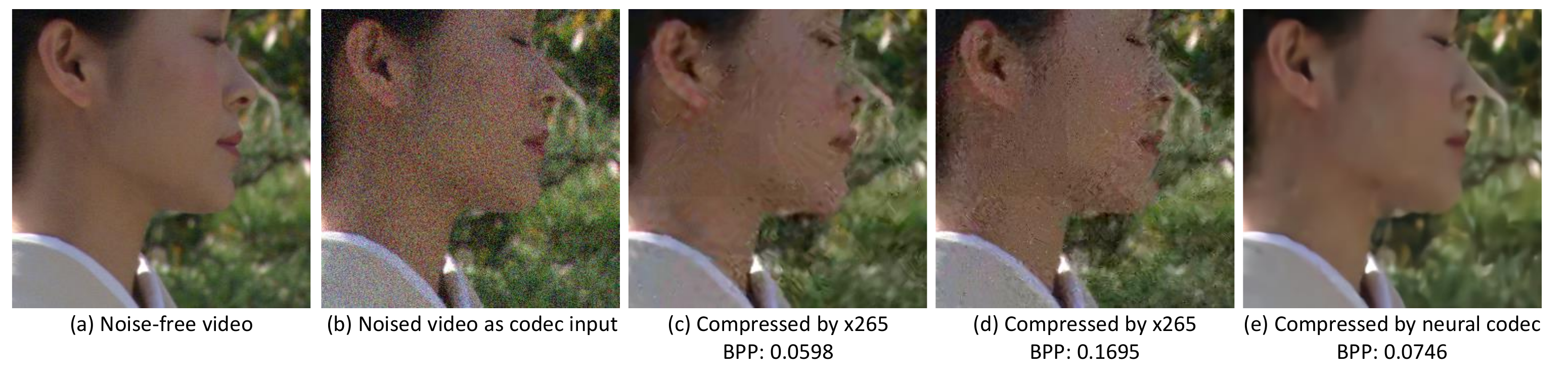}
	   	\vspace{-0.9cm}

		\end{center}
		\caption{Comparison between traidtional codec x265 \cite{FFMPEG} and neural codec \cite{li2021deep} when compressing the noised video (additive white Gaussian noise with $\sigma = 20$). BPP denotes bits per pixel, which measures bitrate cost. 
		}
		\label{Motivation_fig}
		\vspace{-4mm}
	\end{figure*}
	
\section{Related Work} \label{related}

\subsection{Video Restoration}
Existing video restoration methods leveraging temporal correlation can be divided into two categories: the sliding window-based methods and the recurrent methods.

The sliding window-based methods take several adjacent frames as input for each frame. Some methods~\cite{vnlnet, fastdvd} do not depend on explicit motion alignments. VNLNet~\cite{vnlnet} uses the non-local module to search similar patches across frames. FastDVDNet~\cite{fastdvd} uses stacked U-Net~\cite{unet} to progressively fuse the unaligned neighboring frames. By contrast,
ToFlow~\cite{tof} and DVDNet~\cite{dvdnet} use a motion estimation component to explicitly align neighboring frames. 
To explore more temporal correlation, RViDeNet~\cite{crvd} and EDVR~\cite{edvr} propose
feature domain alignment. They align the features of neighboring frames rather than raw pixels
and this mechanism is adopted by most recent methods.

The sliding window-based methods suffer from a narrow temporal scope and cannot leverage the information outside the sliding window. By contrast, the recurrent methods learn the temporal features within a long temporal range, and achieve better performance. EMVD~\cite{emmvd} recurrently combines all past frames as auxiliary information.
Yan \textit{et al.} \cite{yan2019frame} proposed a recurrent feature propagation framework without explicit alignment.
The feature propagation in BasicVSR~\cite{basicvsr} uses the explicit alignment. Recently, BasicVSR++~\cite{
basicvsrpp} achieves excellent performance by using a second-order grid propagation structure and a flow-guided deformable alignment module.

\subsection{Video Compression}

Traditional video codecs, e.g, H.264 and H.265, adopt the hybrid framework which consists of prediction, transform, quantization, entropy coding, and loop-filter. Benefiting from the progress of neural image compression \cite{balle2017,balle2018variational,minnen2018joint}, neural video compression \cite{lu2019dvc,agustsson2020scale,lin2020m,liu2020conditional,li2021deep} recently also has a great development. For example, Lu \textit{ et al.} \cite{lu2019dvc} designed the DVC model, which follows the framework of traditional video codec but uses neural networks to implement all modules therein. Following DVC, Agustsson \textit{ et al.} \cite{agustsson2020scale} designed a more advanced optical flow estimation in scale space. Recently Li \textit{ et al.} \cite{li2021deep} proposed a conditional coding-based framework which achieves better performance.

\section{Motivation}\label{motivation_section}

Our motivation comes from that video compression can filter the noise. Video compression aims at using the least bitrate cost to represent the video.   For traditional codec, the residuals of noisy contents are usually large as they are hard to predict from  reference frames. These residuals contains lots of high-frequency and will consume many bits. To achieve the bitrate saving, traditional codec uses the quantization to discard the residuals of noisy contents, especially for the high-frequency therein, which is like a low-pass filter. We use the traditional codec x265 \cite{FFMPEG}  to conduct an analysis experiment, as shown in Fig.~\ref{Motivation_fig}.
From Fig. \ref{Motivation_fig} (c), we find that the traditional codec x265 can filter the noise in a large degree. Fig. \ref{Motivation_fig} (d) shows, when allocated more bits, x265 will encode the noise but in a much smoother way. 

Different from traditional codec using linear DCT (discrete cosine transform), neural codec will learn a neural encoder to transform video from pixel domain to latent feature domain. The latent feature is then quantized, and its distribution is estimated to perform arithmetic coding.  The distribution is  predicted  more accurately, more bitrate saving is achieved. However, the distributions of  the noisy and uncorrelated  contents are hard to predict well. Thus, to save the bitrate,
these contents are more inclined to be discarded guided by the cross-entropy loss.  Fig. \ref{Motivation_fig} (e) shows the effectiveness of neural codec \cite{li2021deep} (model weights are provided by authors of \cite{li2021deep}). In particular, the neural codec can much better remove the noise therein and keep more semantic information when compared with 
x265.

Inspired by this analysis, we propose utilizing a neural codec to help video restoration. The neural codec is used to filter the noisy information in features via the quantization. 
If the quantization step and data distribution are properly learned, 
 the representations of the noise-perturbed data will be mapped to the same quantized representation with the clean data with high probability.
The noise-robust feature representations will improve the final restoration quality. 
 
Another advantage of using neural codec rather than traditional codec is that the neural codec can be end-to-end trainable and will have better performance when jointly trained with other restoration modules.

\begin{figure*}
		\begin{center}
			\includegraphics[width=\linewidth]{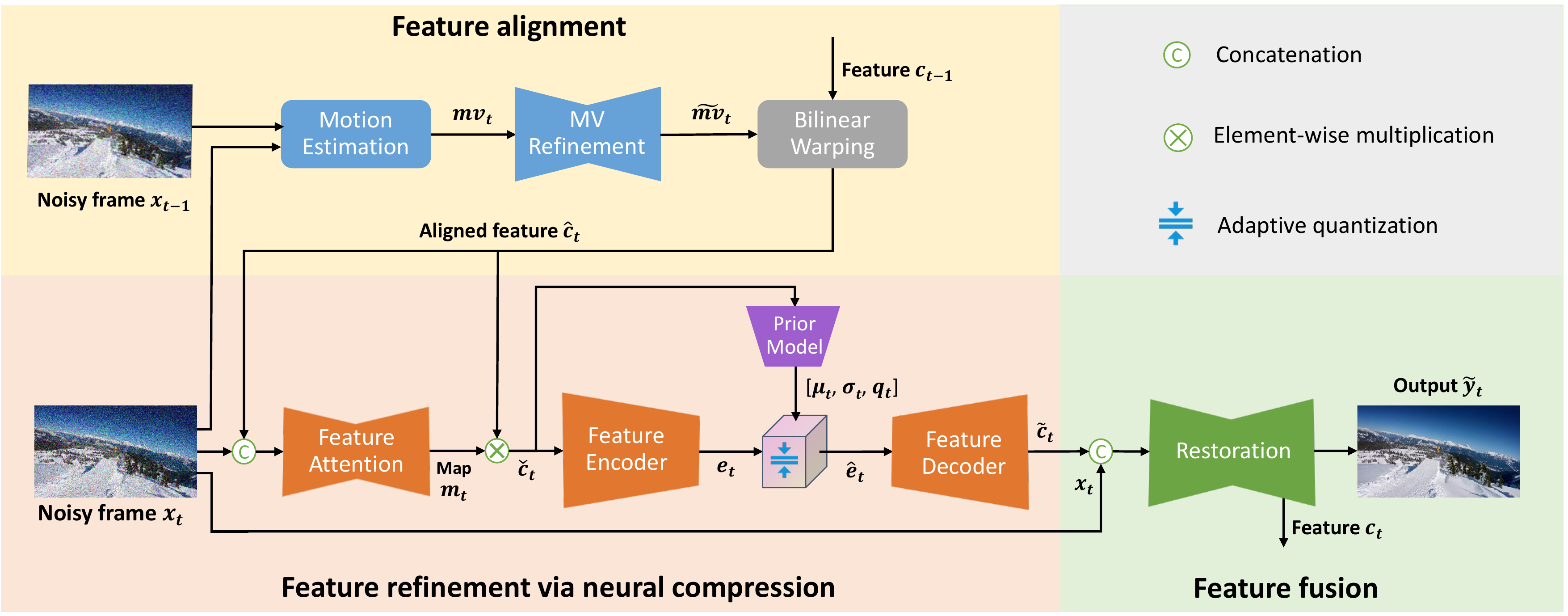}
		\end{center}
		\vspace{-0.4cm}
		\caption{The overall framework of our method. The detailed network structures of each module can be found in appendices. 
		}
		\vspace{-3mm}
		\label{framework}
		
	\end{figure*}

\section{Proposed Method}\label{method}
\subsection{Framework Overview}
We design a neural compression-based framework for video restoration. 
Our framework contains three parts: feature alignment, feature refinement for learning noise-robust feature representations, and feature fusion. The framework is illustrated in Fig.~\ref{framework}.

\textbf{Feature alignment.} Given the noisy frames $x_{t-1}$ and $x_t$, we first use motion estimation to estimate the motion vector (MV) $mv_{t}$. Then we design an MV refinement module to improve the discontinuous MV $mv_{t}$ estimated from noisy video. With the refined MV $\widetilde{mv}_{t}$, the coarse features $\hat{c}_{t}$ are obtained via a bilinear warping function.

 \textbf{Feature refinement.} 
As $\hat{c}_{t}$ contains some noisy and uncorrelated  information, we propose a neural compression-based feature refinement to purify the features. It is noted that, our feature refinement part consists of two modules. One is the attention module and the other is the neural compression module used for noise-robust feature learning.
 
 \textbf{Feature fusion.} 
With the noise-robust features $\tilde{c}_{t}$ and the current frame $x_t$, 
the final output frame $y_t$ is generated through the restoration module. 
Besides $y_t$, the restoration module part will also generate the temporal features $c_t$ used for next step.

\begin{figure}[t]
	   \includegraphics[width=\linewidth]{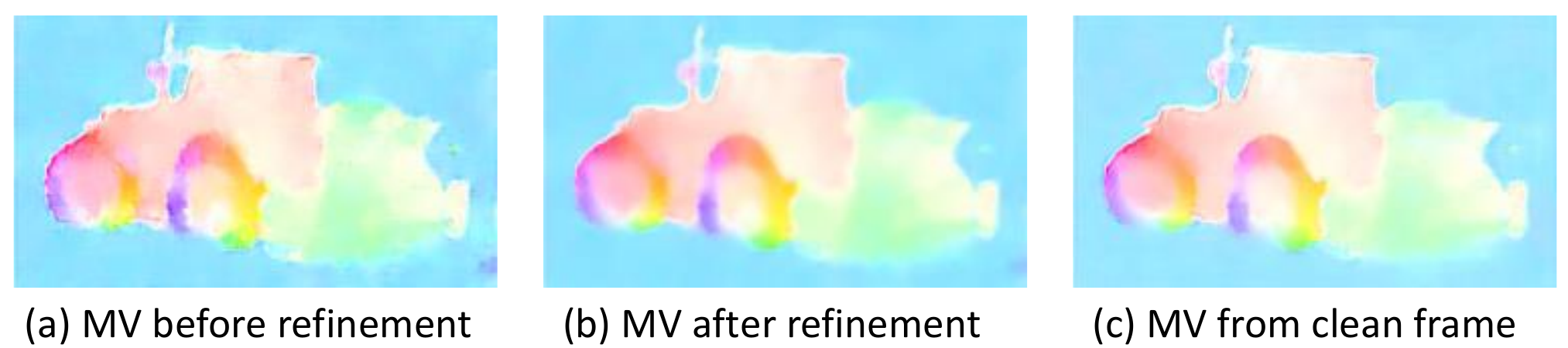}
		
		\caption{Example of MV comparison. We find MV is more accurate and more similar to the MV from clean frames after ME refinement. Zoom in for a better view.}
		\vspace{-3mm}
		\label{vis_flow}
\end{figure}

\subsection{Feature Alignment}\label{mvae}
To align the temporal features from last step to the current frame, we need to predict the MV. In our paper, we use the pre-trained optical flow estimation network SPyNet~\cite{spynet} as our motion estimation module. 

However, estimating accurate MV from the degraded frames is quite difficult. As shown in Fig.~\ref{vis_flow} (a), the MV without any processing suffers from corruption and discontinuity, which is not accurate compared with the MV estimated from clean frames in Fig.~\ref{vis_flow} (c). 
To solve this problem, we propose using an MV refinement module to improve the MV. The MV refinement module adopts a lightweight auto-encoder structure. It encodes the corrupted MV into compact representations and then decodes them to the refined MV. The detailed network structure can be found in appendices. As Fig.~\ref{vis_flow} (b) shows, with our MV refinement, the MV is cleaner and more similar to the MV from the clean frames. 

\subsection{Feature Refinement via Neural Compression}\label{cr}

Previous recurrent methods directly fuse the current frame and the aligned temporal features without any refinement. Actually, the temporal features may still contain some noisy and uncorrelated  information, which disturbs the restoration of the current frame. 

To solve this problem, we propose a feature refinement process to learn noise-robust feature representations. This process consists of two modules, i.e. attention module and neural compression module. For the attention mechanism, many papers \cite{hu2018squeeze,ZhangLLZF19,mei2020pyramid,niu2020single} have studied it and proved its effectiveness. Thus, we design an attention module to scale the temporal features to help the feature learning. To achieve a good trade-off between performance and complexity, we design an auto-encoder-based attention network whose detailed network structure can be found in appendices. 

After the attention module, the temporal features $ \check{c}_{t}$ will be purified by the proposed neural compression module. Following the design in neural image/video compression \cite{balle2018variational,lu2019dvc,li2021deep}, our neural compression module consists of feature encoder-decoder, quantization process, and a prior model. 

First, the temporal features $ \check{c}_{t}$ are encoded to be compact latent codes $e_t$ through the feature encoder:
\begin{equation}
	e_t = Encoder(\check{c}_{t}).
\end{equation}
To achieve the robustness to noise, the quantization is applied to $e_t$. The $e_t\in[s_k,s_{k+1})$ is quantized to value $\frac{s_k+s_{k+1}}{2}$, where $s_k$ and $s_{k+1}$ indicate the numerical range. 
 Let $\ddot{c}_{t}=\check{c}_{t}+\epsilon$ be the noisy feature with noise $\epsilon$. Under the assumption that $Encoder$ is Lipschitz continuous, $\ddot{e}_{t}=Encoder(\ddot{c}_{t})$ and $e_t=Encoder(\check{c}_{t})$ will be located in the same region $[s_k,s_{k+1})$ with high probability if the quantization step   $s_{k+1}-s_k$ is relatively large, then they have the same quantized value. That means, the quantized representation is robust to noisy input. However, the robustness is determined by a prerequisite that the data distribution and quantization step   are properly learned.

\begin{figure}[t]
 		\begin{center}
 			\includegraphics[width=\linewidth]{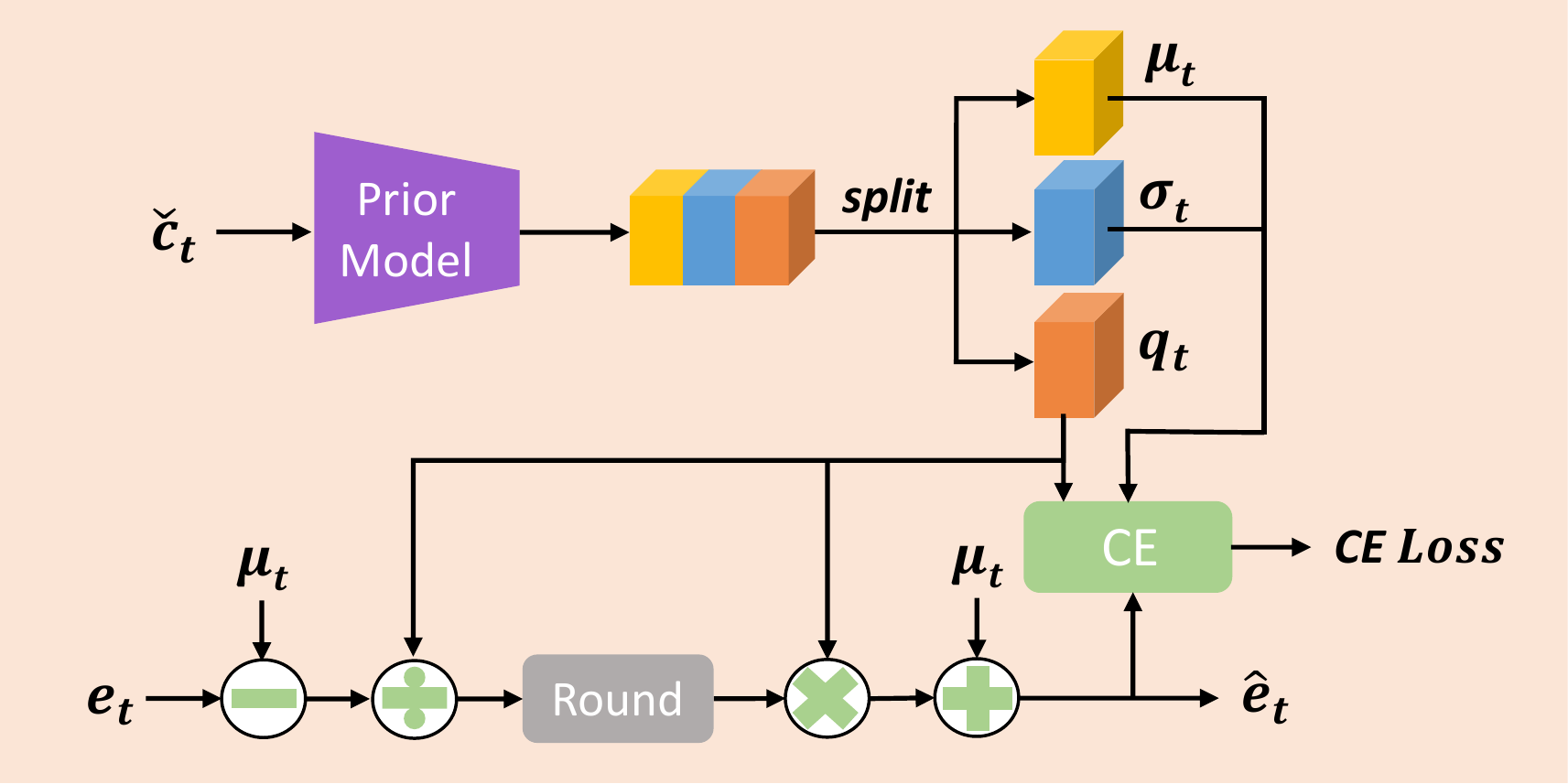}
 			\vspace{-8mm}
 		\end{center}
 	\caption{The illustration of adaptive quantization mechanism. CE means cross-entropy.
 		}
 	\vspace{-4mm}
 		\label{prior}
 	
 	\end{figure}

 Most existing quantization solutions in neural image/video compression only use fixed quantization step. Actually, the content characteristics spatially vary in a large degree. 
A fixed quantization step cannot handle various and complex contents well. For example, a fixed small quantization step fails to remove the noisy  information. A fixed large quantization step instead causes large information loss (i.e., the intrinsic quantization noise). 
Thus, we propose an adaptive quantization mechanism, where the quantization step is learned. The illustration is shown in Fig. \ref{prior}.
First, $e_t$ are divided by the learned quantization step $q_t$ after subtracting the learned mean value $\mu_{t}$. The quotients are then rounded to the closest integers. At last, the quantized latent codes $\hat{e}_t$ are obtained via the opposite operations. The formulation is:
\begin{equation}\label{quant}
   \hat{e}_t = \lceil \frac{e_t-\mu_{t}}{q_t} \rfloor * q_t+\mu_{t}.
\end{equation}
 $\lceil \cdot \rfloor$ is the integer rounding operation. 
 With the quantized latent codes $\hat{e}_t$, the noise-robust temporal features $\tilde{c}_{t}$ are then decoded through the feature decoder:
\begin{equation}
	\tilde{c}_{t} = Decoder( \hat{e}_t ).
\end{equation}

As aforementioned, the data distribution and quantization step need to be properly learned to achieve noise robustness. In practice, we do not know the data distribution, thus we use the prior model to estimate it, and then use the cross-entropy loss to guide the learning of data distribution and quantization step. The  cross-entropy loss is  formulated as:
\begin{equation}\label{entropy_loss}
	\mathcal{L}oss_{CE} = E_{\hat{e}_t} [-log_2 p_{\hat{e}_t}(\hat{e}_t)],
\end{equation}
where $p_{\hat{e}_t}(\hat{e}_t) $ is the estimated probability mass function of the latent codes $\hat{e}_t$. 
 In this paper, we follow ~\cite{li2021deep,pyt} and assume that $p_{\hat{e}_t}(\hat{e}_t)$ follows the Laplace distribution.
The prior model composed by neural network is used for estimating the distribution parameters. Detailed structures of the prior model can be found in appendices. But different from \cite{li2021deep,pyt} that only estimate the distribution parameters $(\mu_t,\sigma_t)$, our prior model also learns the quantization step size $q_t$. With $(\mu_t,\sigma_t,q_t)$, 
the probability estimation of $p_{\hat{e}_t}(\hat{e}_t)$ is calculated as:
\begin{equation}\label{eqa_prior}
   	p_{\hat{e}_t}(\hat{e}_t) = \prod_i( \mathcal{L} (\mu_{t,i},\sigma^2_{t,i}) * \mathcal{U}(-\frac{q_{t,i}}{2},\frac{q_{t,i}}{2})) (\hat{e}_{t,i}), 
\end{equation}
where $i$ specifies the spatial position of each element in $\hat{e}_{t}$. 
According the probability mass function in Eq. \ref{eqa_prior}, we can calculate the cross-entropy loss via Eq. \ref{entropy_loss}. 
The cross-entropy loss guides the compression module to learn proper data distribution and quantization step, and then achieves the robustness to noise. 

\begin{figure}[t]
   \centering
		\begin{center}
 			\includegraphics[width=\linewidth]{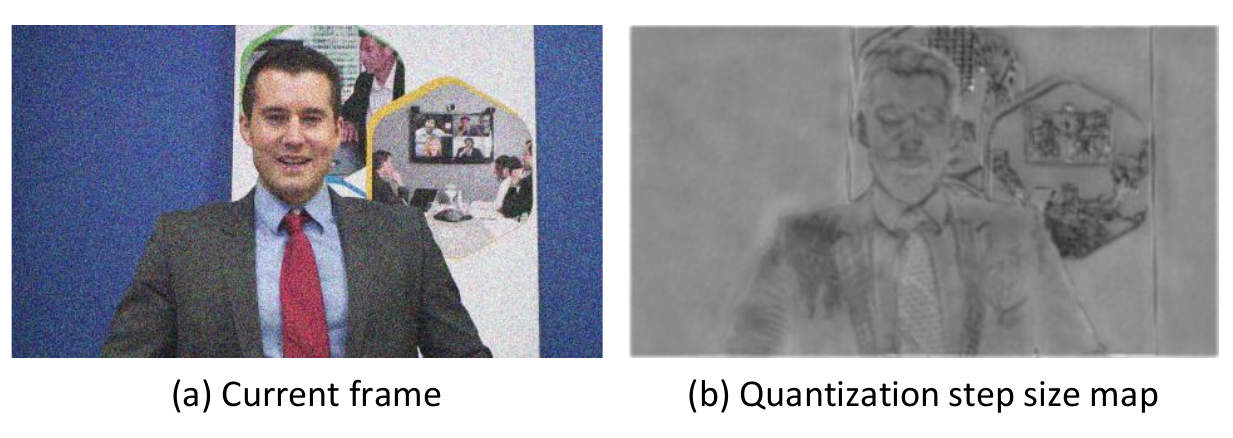}
 			\vspace{-10mm}
 		\end{center}
		\caption{Visualization example of the input frame and the corresponding learned quantization step size map. The frame is  from \textit{Johnny}, HEVC Class E dataset \cite{bossen2013common}.}
		\vspace{-4mm}
		\label{step_map}
		
	   \centering
\end{figure}

In our framework, the quantization step $q_t$ is learnable at spatial-channel-wise. It can be adaptive to regions with different content characteristics. We visualize one channel example of quantization step map in Fig.~\ref{step_map}. The pixel intensity represents the size of quantization step. The larger pixel intensity represents the more noisy information that should be eliminated. As Fig.~\ref{step_map} shows, the quantization step size of the smooth region is usually larger because the noisy information therein is easier to remove. By contrast, the quantization step size of texture region (e.g., the table in the background poster, and there actually exist many details)  is usually smaller.
\subsection{Feature Fusion}\label{fusion}
The feature fusion part contains a restoration module. It will fuse the noise-robust temporal features $\tilde{c}_{t}$ with the current frame $x_t$, and then generates the final output frame $~\tilde{y_t}$. It is noted that, besides final output frame $~\tilde{y_t}$, the restoration module also generates the temporal features $c_t$ used for the next step,  like \cite{basicvsr}. Our restoration module is based on a lightweight U-Net~\cite{unet}. The detailed network structures can be found in appendices.

	\begin{table*}[t]
		\centering
		\scalebox{0.70}{
		\begin{tabular}{cccc ccc ccccc}
			\Xhline{1.0pt}
			 \textbf{$\sigma$} & \textbf{VNLnet~\cite{vnlnet}} & \textbf{DVDNet~\cite{dvdnet}} & \textbf{FastDVDNet~\cite{fastdvd}} & \textbf{EMVD-L~\cite{emmvd}} & \textbf{EMVD-S~\cite{emmvd}} & \textbf{EDVR~\cite{edvr}} & \textbf{BasicVSR~\cite{basicvsr}} & \textbf{BasicVSR++~\cite{basicvsrpp}} & \textbf{Ours}\\
			\Xhline{1.0pt}
			10    & 37.10/0.9637 & 36.08/0.9592 & 36.44/0.9624 & 36.56/0.9624 & 35.01/0.9442 & 37.16/0.9658 & 37.12/0.9674 & \textcolor{red}{\textbf{37.27}}/\textcolor{blue}{\textbf{/0.9682}}& \textcolor{blue}{\textbf{37.17}}/\textcolor{red}{\textbf{0.9684}} &\\ 
			20    & 33.88/0.9360 & 33.49/0.9307 & 33.43/0.9334 & 33.27/0.9320 & 31.65/0.8927 & 34.09/0.9379 & 34.13/0.9397 & \textcolor{red}{\textbf{34.25}}/\textcolor{blue}{\textbf{0.9411}} & \textcolor{blue}{\textbf{34.22}}\textcolor{red}{\textbf{/0.9437}}&\\ 
			30   & 31.95/0.9096      & 31.79/0.9023 & 31.68/0.9066 & 31.40/0.9032 & 29.94/0.8678 & 32.31/0.9125 & 32.33/0.9157 &
			\textcolor{blue}{\textbf{32.55}}/\textcolor{blue}{\textbf{0.9168}}&\textcolor{red}{\textbf{32.57/0.9184}}& \\ 
		   40   & 30.55/0.8814 & 30.55/0.8745 & 30.46/0.8812 & 30.05/0.8761 & 28.64/0.8328 & 31.02/0.8887 & 31.05/0.8929 &
		   \textcolor{blue}{\textbf{31.28}}/\textcolor{blue}{\textbf{0.8936}} &\textcolor{red}{\textbf{31.39/0.8970}}& \\ 
			50   & 29.47/0.8561   &   29.56/0.8480 & 29.53/0.8573 & 29.15/0.8528 & 27.83/0.8082 & 30.06/0.8660 & 30.11/0.8690   &\textcolor{blue}{\textbf{30.32}}/\textcolor{blue}{\textbf{0.8696}} & \textcolor{red}{\textbf{30.45/0.8770}}& \\\hline
			FLOPs (G) & - & - & 665 & 1106 & 5 & 3089 & 2947 & 3402 & 771 \\ 
			\hline
			\Xhline{1.0pt}
		\end{tabular}}
		\vspace{-2mm}
		\caption{PSNR/SSIM comparison with SOTA video denoising methods on synthetic dataset Set8. The best performance is highlighted in \textcolor{red}{\textbf{red}} (1st best) and \textcolor{blue}{\textbf{blue}} (2nd best). Our method achieves the best SSIM on all noise levels.} \label{davis}
		\vspace{-2mm}
	\end{table*}

		\begin{table*}[t]
		\centering
		\scalebox{0.74}{
			\begin{tabular}{ccc ccc ccc c}
			\Xhline{1.0pt}
			 & \textbf{FastDVDNet~\cite{fastdvd}} & \textbf{EDVR~\cite{edvr}} & \textbf{RViDeNet~\cite{crvd}} &   \textbf{EMVD-L~\cite{emmvd}} & \textbf{EMVD-S~\cite{emmvd}} & \textbf{BasicVSR~\cite{basicvsr}} & \textbf{BasicVSR++~\cite{basicvsrpp}} & \textbf{Ours} &\textbf{Ours-L} \\
			\Xhline{1.0pt}
			PSNR &44.30& 44.71 & 44.08 &      44.48 & 42.63 & 44.80 & \textcolor{blue}{\textbf{44.98}} & 44.72   & \textcolor{red}{\textbf{45.09}} \\ 
		   SSIM &0.9881 & 0.9902   & 0.9881   & 0.9895 & 0.9851 & 0.9903 & 0.9903 & \textcolor{blue}{\textbf{0.9906}} & \textcolor{red}{\textbf{0.9909}} \\ 
		   Runtime (ms) & 132 & 1511 & 1254 & 246 & 59 & 425 & 488 & 188 & 275 \\\hline

			\hline
			\Xhline{1.0pt}
		\end{tabular}}
        \vspace{-2mm}
		\caption{Comparison with SOTA video denoising methods on real-world dataset CRVD~\cite{crvd}.  Our method with default setting outperforms other fast methods  and gets close to the slow method. With a more powerful restoration module (i.e. 'Ours-L'), we can achieve SOTA performance in terms of both PSNR and SSIM.   The runtime is the average frame runtime for the whole dataset on a P100 GPU. } \label{crvd}
		\vspace{-5mm}
	\end{table*}

	\begin{table*}[t]
		\centering
		\scalebox{0.69}{
			\begin{tabular}{lcc ccc ccccc}
			\Xhline{1.0pt}
			 & & \textbf{MS-CSC~\cite{mscsc}} & \textbf{SE~\cite{se}} & \textbf{SpacCNN~\cite{spaccnn}} & \textbf{FastDerain~\cite{fastderain}} & \textbf{J4RNet-P~\cite{j4r} } & \textbf{FCRVD~\cite{fcrvd}} &\textbf{RMFD~\cite{rfmd}} & \textbf{BasicVSR++~\cite{basicvsrpp}} & \textbf{Ours} \\
			\Xhline{1.0pt}
			\multirow{2}*{RainSynAll100} & PSNR &    16.19 & 15.29 & 18.39 & 17.09 & 19.26& 21.06 & 25.14 & \textcolor{blue}{\textbf{27.67}} &\textcolor{red}{\textbf{28.11}} \\ 
		   & SSIM &   0.5078   & 0.5053 & 0.6469 & 0.5824 & 0.6238 & 0.7405 & \textcolor{blue}{\textbf{0.9172}} & 0.9135 & \textcolor{red}{\textbf{0.9235}} \\ \hline
		   
		   \multirow{2}*{RainSynComplex25} & PSNR &    16.96 & 16.76 & 21.21 & 19.25 & 24.13& 27.72 & 32.70 & \textcolor{blue}{\textbf{33.42}} &\textcolor{red}{\textbf{34.27}} \\ 
		   & SSIM &   0.5049   & 0.5273 & 0.5854 & 0.5385 & 0.7163 & 0.8239 & 0.9357 & \textcolor{blue}{\textbf{0.9365}} & \textcolor{red}{\textbf{0.9434}} \\ \hline

			\hline
			\Xhline{1.0pt}
		\end{tabular}}
        \vspace{-2mm}
		\caption{Comparison with SOTA video deraining methods on RainSynComplex25~\cite{j4r} and RainSynAll100~\cite{rfmd}. We train the BasicVSR++~\cite{basicvsrpp} using the same setting as ours. Other baseline results are provided by RMFD~\cite{rfmd} paper.} \label{derain}
		\vspace{-2mm}
	\end{table*}
	
		\begin{table*}[t]
		\centering
		\scalebox{0.74}{
			\scalebox{0.97}{\begin{tabular}{ccc ccc ccc ccc}
			\Xhline{1.0pt}
			 & \textbf{DCP~\cite{dcp}} & \textbf{GDNet~\cite{gridn}} & \textbf{DuRN~\cite{durn}} &   \textbf{KDDN~\cite{kddn}} & \textbf{MSBDN~\cite{mbsdn}} & \textbf{FFA~\cite{ffa}} & \textbf{VDN~\cite{vdn} } & \textbf{EDVR~\cite{edvr}} &\textbf{CG-IDN~\cite{cgidn}} & \textbf{BasicVSR++~\cite{basicvsrpp}} & \textbf{Ours} \\
			\Xhline{1.0pt}
			PSNR &11.03& 19.69 & 18.51 &    16.32 & 22.01 & 16.65 & 16.64 & 21.22 & \textcolor{blue}{\textbf{23.21}} & 21.68 & \textcolor{red}{\textbf{23.63}} \\ 
		   SSIM &0.7285 & 0.8545 & 0.8272 &   0.7731 & 0.8759   & 0.8133 & 0.8133 & 0.8707 & \textcolor{blue}{\textbf{0.8836}} & 0.8726 & \textcolor{red}{\textbf{0.8925}} \\ \hline

			\hline
			\Xhline{1.0pt}
		\end{tabular}}}
        \vspace{-2mm}
		\caption{Comparison with SOTA video dehazing methods on REVIDE~\cite{cgidn} testset. We train the BasicVSR++~\cite{basicvsrpp} using the same setting as ours. Other baseline results are provided by CG-IDN~\cite{cgidn} paper.} \label{dehaze}
		\vspace{-5mm}
	\end{table*}

\subsection{Loss Function}
In our method, the loss function includes two items:
\begin{equation}
		\mathcal{L}oss= \sum_{t=1}^{n} \mathcal{L}oss_{\mathcal{L}2}(y_t, ~\tilde{y_t}) + \lambda \cdot \mathcal{L}oss_{CE}(\hat{e}_t).
\end{equation}
 $y_t$ and $~\tilde{y_t}$ are the clean and estimated frames, respectively.
$ \mathcal{L}oss_{\mathcal{L}2}$ is the L2 loss  and $\mathcal{L}oss_{CE}$ is the cross-entropy loss. 
 To  learn noise-robust feature representation and then let it help the final  reconstruction,  we adopt a  two-stage training scheme and the details are in appendices.

\section{Experiment}~\label{experiment}
We evaluate our method on several video restoration tasks, including denoising, deraining, and dehazing. 

\subsection{Dataset}
\textbf{Video denoising.} Both synthetic dataset and real-world dataset are tested. For the synthetic dataset, we follow the setting in FastDVDNet~\cite{fastdvd}. DAVIS2017 train-val set containing 90 videos is used for training. Set8 is used for testing. We add the additive white Gaussian noise (AWGN) to the clean video to synthesize the noisy video. Five noise levels, i.e. $\sigma$=10, 20, 30, 40, 50 are tested. For the real-world dataset, we follow the setting in EMVD~\cite{emmvd} and use the dataset from RViDeNet~\cite{crvd}. 
It consists of a captured raw video dataset (CRVD) and a synthetic raw video dataset (SRVD). 
Following EMVD and RviDeNet, we use CRVD scene 1$\sim$6 plus SRVD for training and CRVD scene 7$\sim$11 for testing.

\textbf{Video deraining.} Following ~\cite{rfmd}, we test our method on RainSynComplex25~\cite{j4r} and RainSynAll100~\cite{rfmd} datasets. RainSynComplex25 contains 190 videos for training and 25 videos for testing.
RainSynAll100 contains 900 videos for training and 100 videos for testing.

\textbf{Video dehazing.} We use REVIDE~\cite{cgidn} dataset which captures the pairs of hazy and corresponding haze-free videos in the
same scene by an acquisition system. It contains 42 videos for training and 6 videos for testing.

\subsection{Result on Video Denoising }
We compare our method with these baselines: VNLNet~\cite{vnlnet}, DVDNet~\cite{dvdnet}, FastDVDNet~\cite{fastdvd}, EMVD~\cite{emmvd}, EDVR~\cite{edvr}, BasicVSR~\cite{basicvsr}, BasicVSR++~\cite{basicvsrpp}, and RViDeNet~\cite{crvd}.
EMVD has several network structure configurations with different complexities. The large (EDVR-L) and small (EMVD-S) models are tested (more details about configurations are in appendices). The original BasicVSR/BasicVSR++ are bi-directional methods that leverage the temporal features from both the future and the past frames. To compare with other methods more fairly, we modified BasicVSR/BasicVSR++ to uni-directional methods that only use the temporal features from the past frames.

\textbf{Quantitative Comparison.}
We use peak signal-to-noise ratio (PSNR) and structural similarity index measure (SSIM) as quantitative evaluation metrics. 
We present the results on synthetic noise video in Table~\ref{davis} and the results on real-world noisy video in Table~\ref{crvd}. 
For the synthetic video, as Table~\ref{davis} shows, our method achieves the best SSIM on all noise levels. For PSNR, our method outperforms the second-best method BasicVSR++~\cite{basicvsrpp} with as least 0.11dB gain when the noise level is high ($\sigma$ = 40 or 50). 
In addition, we   find   the quality improvement over BasicVSR++ is larger when the noise level is higher. It verifies that our proposed neural compression module can effectively filter the noise.
It is also worth noting that the FLOPs of our method is only 0.23 times of BasicVSR++, which shows that our method achieves a much better trade-off between quality and complexity.
Compared with the low-complexity methods FastDVDNet~\cite{fastdvd} and EMVD-L~\cite{emmvd}, our method achieves significant quality improvement.
For the real-world noisy video, it should be admitted that our method with default setting currently cannot outperform BasicVSR and BasicVSR++ in terms of PSNR but is better than them in terms of SSIM. When compared with low-complexity method FastDVDNet and EMVD-L, our method can achieve the best quality. Besides, if we change our U-Net-like~\cite{unet} restoration network to a W-Net-like~\cite{wnet} restoration network with more complexity (more details are in appendices), denoted as 'Our-L' in Table~\ref{crvd}, we can achieve the best PSNR and SSIM at the same time, but the complexity is still much less than that of BasicVSR and BasicVSR++.

\begin{figure*}[ht]
	\begin{center}
		\includegraphics[width=1.0\linewidth]{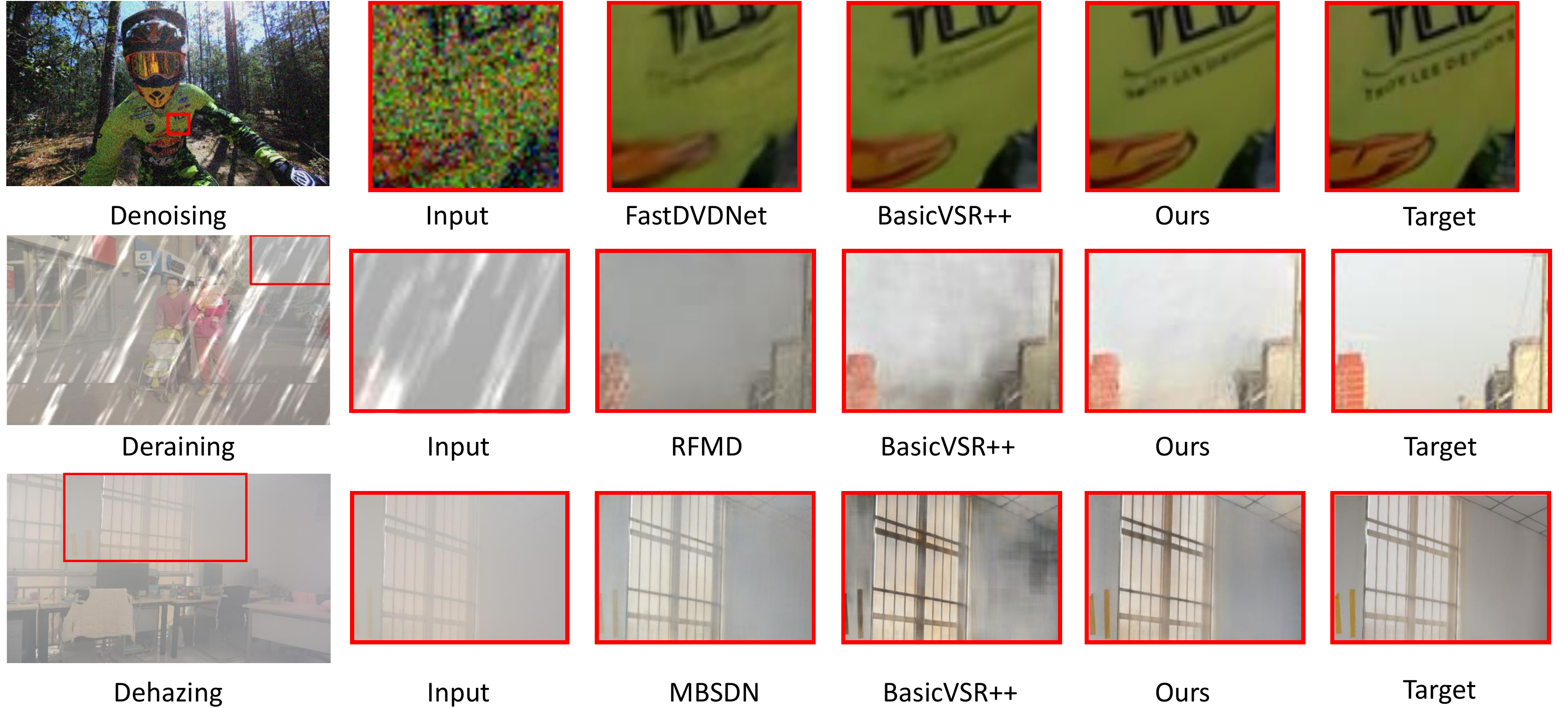}
	\end{center}
	\vspace{-6mm}
	\caption{Denoising: \textit{motorbike} video from  Set8 testset with noise variance 50. Deraining:   \textit{0985} video  from RainSynAll100 testset. Dehazing: \textit{L006} video from REVIDE dataset.
}
	\label{fig4}
	\vspace{-3mm}
\end{figure*}

\textbf{Qualitative Comparison.}
Fig.~\ref{fig4} shows the visual quality comparison.
As Fig.~\ref{fig4} shows,  FastDVDNet without feature alignment suffers from serious distortion in the text region.  The results of  BasicVSR++ are quite blurry caused by the propagated noise in temporal features. 
By contrast, our neural compression-based method can learn noise-robust features and is able to restore much clearer textures.
More visual comparisons are in appendices.

		\begin{table}[t]
		\centering
		
		\scalebox{0.92}{\begin{tabular}{l|cccccc}
			\Xhline{1.0pt}
		    & $M_a$ & $M_b$ & $M_c$ & $M_d$ \\
			\hline\hline
			
		    \textbf{MVR} &    & \checkmark            & \checkmark & \checkmark   \\ 
		    \textbf{NCFL} &    &             & \checkmark & \checkmark\\
		    \textbf{FA} &    &                               &   & \checkmark \\\hline
      PSNR   & 29.75 & 29.87 &         30.29   & 30.45 \\
		   
			\Xhline{1.0pt}
		\end{tabular}}
		\caption{The ablation study on different modules. Tested on Set8 ($\sigma = 50$). 
		MVR is the MV refinement, NCFL is the neural  compression-based feature learning. FA is the feature attention. } \label{ablation}
	\end{table}

		\begin{table}[t]
		\centering
		\scalebox{0.92}{
		\begin{tabular}{l|ccc}
			\Xhline{1.0pt}
		    & \textbf{NCFL-AdapQ} & \textbf{NCFL-FixedQ} & \textbf{NCFL-NoQ} \\
			\hline\hline
      PSNR    & 30.29 & 29.86   & 29.98 \\
		   
			\Xhline{1.0pt}
		\end{tabular}}
		\caption{The ablation study on quantization. Tested on Set8 ($\sigma = 50$). 
	      NCFL-AdapQ is our default model with adaptive quantization, i.e. $M_c$ in Table \ref{ablation}. NCFL-FixedQ means we use the fixed quantization step as in existing neural video codecs. NCFL-NoQ removes the quantization and  is only a vanilla auto-encoder. } \label{cae}
	      \vspace{-2mm}
	\end{table}

\subsection{Result on Video Deraining }
We compare our method with prior SOTA video deraining methods, including MS-CSC~\cite{mscsc}, SE~\cite{se},  SpacCNN~\cite{spaccnn}, FastDerain~\cite{fastderain}, J4RNet-P~\cite{j4r}, FCRVD~\cite{fcrvd}, RMFD~\cite{rfmd}, and BasicVSR++~\cite{basicvsrpp}. 
Since RainSynAll100 uses the rain accumulation degradation to generate the rainy video,  parts of the baseline methods including SE, MS-CSC, SpacCNN, and FastDerain could not handle this degradation thus MRF~\cite{mrf} is used as post-processing. More details could be found in~\cite{rfmd}.
FCRVD, RMFD, BasicVSR++, and our method could handle this degradation without extra post-processing. As Table~\ref{derain} shows, BasicVSR++ beats RMFD in terms of PSNR and SSIM on RainSynComplex25 but is slightly worse than RMFD in terms of SSIM on RainSynAll100. By contrast,  with the robust temporal features,  our method achieves the best PSNR and SSIM on both datasets.  Our method brings PSNR gain of 0.44 dB and SSIM gain of 0.0063 on RainSynAll100. We also 
test RainSynLight25~\cite{j4r} and NTURain~\cite{spaccnn}. Their  results are provided in appendices. Fig.~\ref{fig4} also shows the visual quality comparison. We can see that our model could   remove   rain streak well, and produces   clearer and  more visual pleasing results.

\subsection{Result on Real-World Video Dehazing }
Table~\ref{dehaze} shows the comparison between our method and prior SOTA real-world video dehazing method: DCP~\cite{dcp}, GDNet~\cite{gridn}, DuRN~\cite{durn}, KDDN~\cite{kddn}, MSBDN~\cite{mbsdn}, FFA~\cite{ffa}, VDN~\cite{vdn}, EDVR~\cite{edvr}, CG-IDN~\cite{cgidn}, and BasicVSR++~\cite{basicvsrpp}. As Table~\ref{dehaze} shows, BasicVSR++ outperforms EDVR but is worse than MBSDN and CG-IDN that are specially designed for dehazing task. By contrast, our method has 0.42 dB PSNR and 0.0089 SSIM 
improvements when compared with the second-best method CG-IDN.
In addition, the parameters of our method are 16M, and only 0.70 times of CG-IDN that has 23M parameters.
As Fig.~\ref{fig4} shows, the result of our method is more visual pleasing.  

\subsection{Ablation Study}
This paper proposes three key modules: the MV refinement (MVR) for improving MV, the neural compression-based feature learning (NCFL) incorporated with adaptive quantization, and feature attention (FA). We study the effect of these modules and report the results in Table~\ref{ablation}. Without MVR, NCFL, and FA, the baseline model only contains a motion estimation module, bilinear warping, and a restoration module. 

\textbf{MV refinement (MVR).} As Table~\ref{ablation} shows, the baseline model $M_a$ only achieves PSNR 29.75dB. It suffers from the discontinuous MV estimated from noisy video. When enabling our MVR, the MV is refined and $M_b$ reaches PSNR 29.87 dB. Our MVR  brings 0.12 dB PSNR improvement.

\textbf{Neural compression-based feature learning (NCFL).} If we further combine NCFL and MVR, $M_c$ reaches PSNR 30.29 dB and improves 0.42 dB compared with $M_b$. The significant improvement  verifies that the effectiveness of  NCFL. In addition, we also study two variants of NCFL. As Table~\ref{cae} shows, the PSNR of a vanilla auto-encoder without quantization (i.e. NCFL-NoQ) drops to 29.98dB. 
This denotes that the improvement brought by NCFL mainly comes from the adaptive quantization mechanism rather than the increase of model parameters. In addition, we also test NCFL-FixedQ where a fixed quantization step is used as many existing neural video codecs do. The PSNR of NCFL-FixedQ drops to 29.89 dB. Its performance is even worse than NCFL-NoQ. This shows that a fixed quantization step instead loses some useful information and fails at learning noise-robust representations. By contrast,
a learnable quantization step at spatial-channel-wise can adaptively filter the noise and purify the temporal features with different content characteristics, which is quite important. 

\textbf{Feature attention (FA).} In this paper, we also propose a FA module to further help the feature learning. As Table \ref{ablation} shows, $M_d$ achieves PSNR 30.45 dB. FA boosts the PNSR by 0.16 dB, which shows its effectiveness.

\subsection{NCFL on Different Degradations}
Table \ref{ablation} and Table \ref{cae} invesitage NCFL under AWGN degradation. 
However, our NCFL is not confined to AWGN. It is also very effective for other complex degradations, such as real-word denoising, deraining, and dehazing.
Table \ref{multiTypes}  shows the comprehensive study on multiple degradations. For example, the comparison between \textit{$M_1$} and \textit{$M_3$}  shows that NCFL can achieve 0.81 dB gain for deraining. These substantial improvements verify the effectiveness of our NCFL. In addition, the comparison between \textit{$M_2$} and \textit{$M_3$} shows that cross-entropy loss can effectively guides the learning of NCFL under multiple degradations.
 
		\begin{table}[t]
		\centering
		\scalebox{0.85}{
			\begin{tabular}{lcccc}
            \Xhline{1.0pt}
             & Deraining &  Dehazing & RWD & AWGN \\\hline
			  \textit{$M_1$}: w/o NCFL  & 27.30 & 23.07 & 44.48 & 29.99   \\ 
			  \textit{$M_2$}: w/ NCFL (w/o CE) & 27.64 & 23.30 & 44.56 & 30.20 \\
			\textit{$M_3$}: w/ NCFL & 28.11 & 23.63 & 44.72 & 30.45 \\ 
			
			\Xhline{1.0pt}
		\end{tabular}}
		\caption{Study on NCFL under different degradation types. CE means the cross-entropy loss. RWD means real-world denoising. AWGN indicates additive white Gaussian noise.  }
		\label{multiTypes}
	\end{table}  

\subsection{Bi-directional Video Denoising}
In previous experiments, we focus on the uni-directional setting where the temporal features only come from the past time. For the bi-directional setting, both the temporal features come from the past time and the future time can be used. One advantage of our method is that our method can be extended to bi-directional  setting easily.
We test the bi-directional model of BasicVSR~\cite{basicvsr}, BasicVSR++~\cite{basicvsrpp}, and our bi-directional model. The PSNR and complexity comparison is shown in Table~\ref{bidirection}. As Table~\ref{bidirection} shows, the bi-directional setting brings BasicVSR 0.59 dB gain, BasicVSR++ 0.78 dB gain, and our method 0.76 dB gain  with about 2x complexity. Under bi-directional setting, our  method still outperforms BasicVSR++ by 0.11 dB with only 0.21x FLOPs.
		\begin{table}[t]
		\centering
        \scalebox{0.90}{
		\begin{tabular}{cccc}
			\Xhline{1.0pt}
		    \textbf{Direction}& \textbf{Method} & \textbf{PSNR} & \textbf{FLOPs (G)}\\
			\hline\hline
			
		Uni-direction & BasicVSR   & 30.11 &   2947          \\
		   Uni-direction & BasicVSR++ & 30.32 &   3402          \\
		   Uni-direction & Ours &   30.45   & 771   \\\hline
		    Bi-direction& BasicVSR &    30.68 &   5855      \\
		    Bi-direction& BasicVSR++ &   31.10 &   7097      \\
          Bi-direction& Ours   & 31.21 &   1522 \\

			\Xhline{1.0pt}
		\end{tabular}}
		\caption{Bidirectional video denoising on Set8 with $\sigma$ = 50. } \label{bidirection}
	\end{table} 

\section{Conclusion and Limitation}~\label{conclusion}
In this paper, we have designed a neural compression-based video restoration framework. We are inspired by the fact that neural video codec can naturally filter the noise, and then propose using neural compression to purify the temporal features and learn noise-robust feature representations. To solve the problem that the fixed quantization step harms the inherent textures, we propose a  learnable quantization mechanism at spatial-channel-wise to achieve robustness to noise.
At the same time, an attention module and an MV refinement module are proposed to further boost the performance. 
 Experimental results show that the proposed method achieves a much better quality-complexity trade-off than previous SOTA methods.
 
Although our method is faster than most previous SOTA methods,  the inference speed of our method still does not meet the requirements of real-time scenarios. In the future, we will continue to improve the efficiency of our method for real-time video restoration. 
 
{\small
	\bibliographystyle{ieee_fullname}
	\bibliography{arxiv_text}
}
\clearpage
\begin{appendices}
	Appendices  includes detailed network architecture, training details for different tasks, and additional ablation studies.  More comparisons on video denoising, video deraining and video dehazing are also presented. 
	
\begin{figure*}[h]
	\begin{center}
		\includegraphics[width=\linewidth]{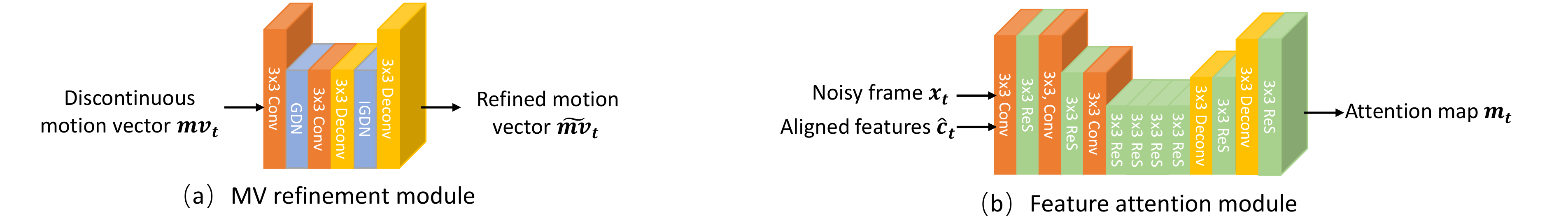}
		\vspace{-0.6cm}
		
	\end{center}
	\caption{The structure of   MV refinement module and  feature attention module. }
	\label{mv}
\end{figure*}

\begin{figure*}[h]
	\begin{center}
		\includegraphics[width=\linewidth]{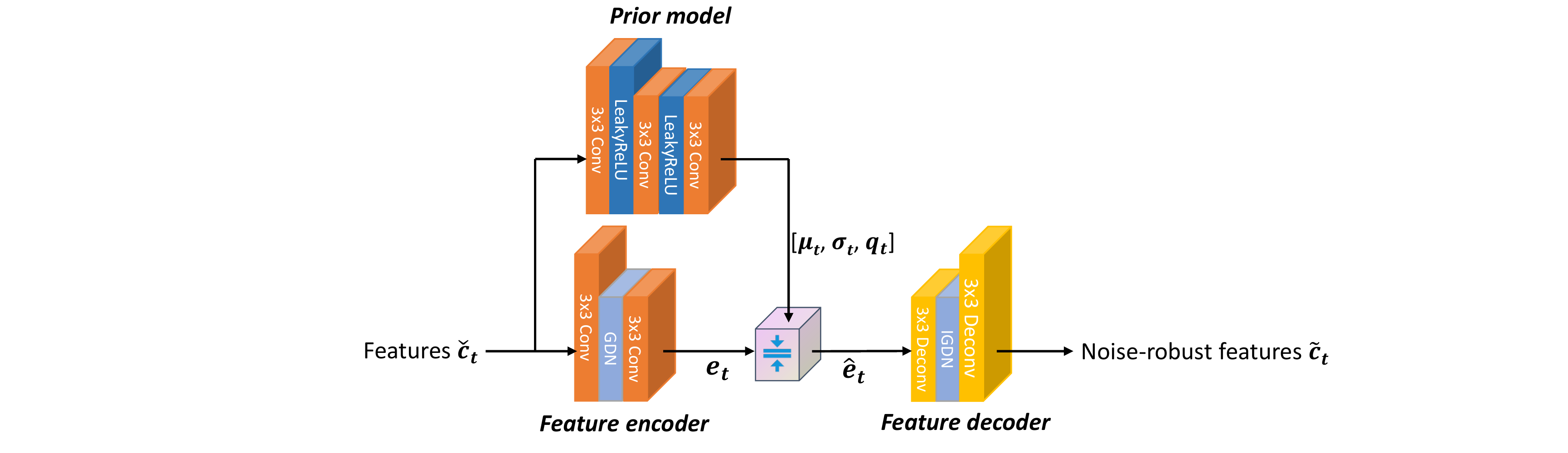}
		\vspace{-0.6cm}
		
	\end{center}
	\caption{The structure of neural compression-based feature  learning module. }
	\label{nccl}
\end{figure*}

\begin{figure*}[h]
	\begin{center}
		\includegraphics[width=\linewidth]{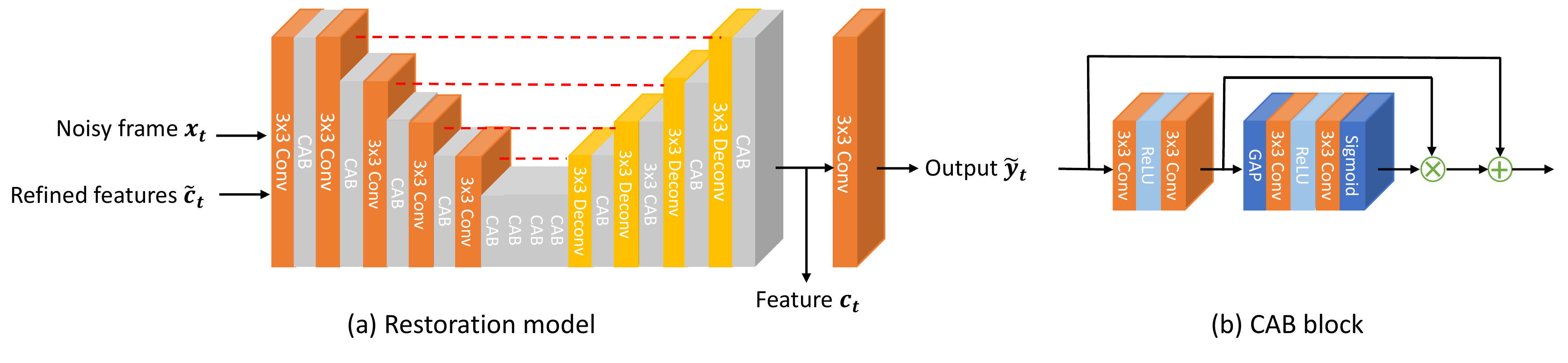}
		\vspace{-0.6cm}
		
	\end{center}
	\caption{The structure of U-Net-like restoration module and CAB block. }
	\label{unet}
\end{figure*}
\begin{figure*}[h]
	\begin{center}
		\includegraphics[width=\linewidth]{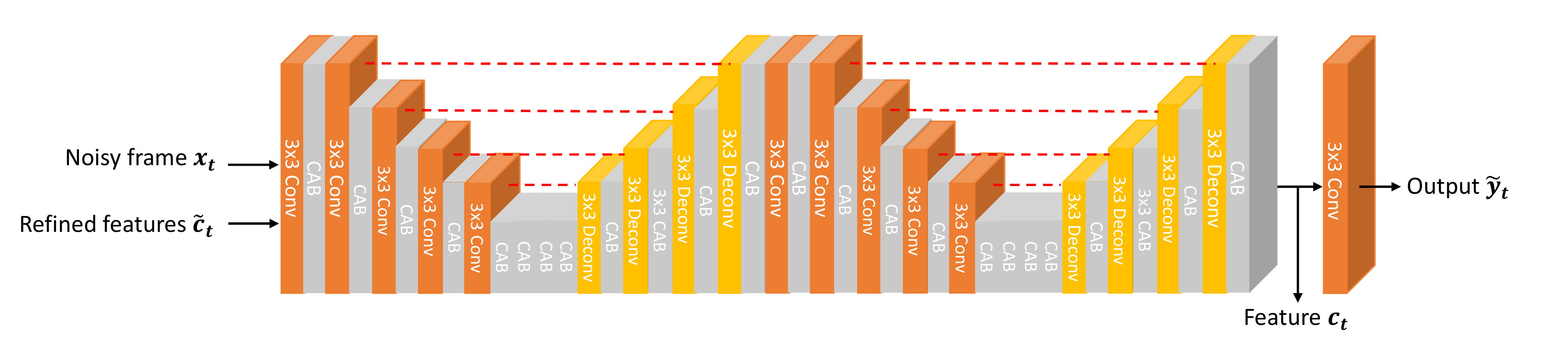}
		\vspace{-0.6cm}
		
	\end{center}
	\caption{The structure of W-Net-like restoration module. }
	\label{wnet}
\end{figure*}

\section{Network Architecture}

Our framework contains three parts: feature alignment, feature refinement (including feature attention module and  neural compression-based feature learning module), and feature fusion. 
In this appendix, we present the network architecture details. 

\noindent\textbf{MV refinement module.}  The structure of the MV refinement module is  shown in Fig.~\ref{mv} (a). The MV refinement module encodes the corrupted MV  into a compact representation and then decodes it to the refined MV. Specifically,  we use two convolutional layers with stride=2 to encode the corrupted MV and two deconv layers to decode the refined MV. The number of the channel of the intermediate features is 64. 
As GDN~\cite{gdn} could reduce the  statistical dependencies of the features and compact  the features,
we use GDN~\cite{gdn} in the encoder and use the inverse-GDN (IGDN)  in the decoder correspondingly. 

\noindent\textbf{Feature attention module.} As Fig.~\ref{mv} (b) shows, the feature attention module takes the noisy frame $x_t$ and aligned features $\hat{c}_{t}$ as input, then generates the spatial-channel attention map $m_t$. 
The  feature attention module is based on an auto-encoder with the following modifications. First,  we use Resblock~\cite{edsr} to extract the features at each scale. Second,  we add 4 Resblocks  at the largest scale to increase the receptive field with minor computation cost. To keep the feature attention module lightweight, the numbers of the channels of the intermediate features are 16, 32, 64 for three scales, respectively. 

\noindent\textbf{Neural compression-based feature learning  module.} As Fig.~\ref{nccl} shows, the neural compression-based feature learning module contains a prior model, a feature encoder, and a feature decoder. The prior model learns to estimate the  parameters $(\mu_t,\sigma_t,q_t)$ that are used in the adaptive quantization and the  $\mathcal{L}oss_{CE}$. 
The feature encoder encodes the  features  $\check{c}_{t}$   into  compact latent codes $e_{t}$. Then the latent codes $e_{t}$ are processed by our proposed adaptive quantization. At last, the refined  features $\tilde{c}_t$ are decoded from the processed latent codes $\hat{e}_t$. For the prior model, we use three convolutional layers (two layers are with stride=2) to estimate the parameters, where LeakyReLU is used as the activation function.
For the feature encoder and decoder, we use two convolutional layers with stride=2 to encode the  features and two deconvolutional layers to decode the  noise-robust  features. The number of the channel of the intermediate features is 64.

\noindent\textbf{Restoration module.} As Fig.~\ref{unet} shows, the restoration module fuses the noise-robust  features $\tilde{c}_{t}$ with the current frame $x_t$, and then generates the final output frame $~\tilde{y}_t$. At the same time, the features $c_{t}$ used by next step are also generated. Following  previous image restoration methods~\cite{mbsdn,sid}, we adopt a deep U-Net~\cite{unet} as our restoration module for synthetic video denoising, video deraining and video dehazing. We also leverage the channel-attention block~\cite{rcan} (CAB)  to extract the features at each scale. The structure of CAB is shown in Fig.~\ref{unet} (b) and the GAP means the global average pooling.  The features from the skip connection are processed by a convolutional layer. The  numbers of the channels  of the intermediate features are 32, 64,  128, 256, 512 for five scales, respectively. 
We also propose a more powerful W-Net-like~\cite{wnet} restoration module for real-world video denoising which is a more challenging task. As Fig.~\ref{wnet} shows, we cascade two U-Net restoration modules as our W-Net-like restoration module.  The  number of the channel of the intermediate features of each scale remains unchanged.

\section{Training Details}
We adopt AdamW optimizer and cosine annealing learning rate scheduler. The initial learning rates of the  motion estimation module and other modules are set to 2.5e-5 and 2e-4, respectively. The weights of the motion estimation module are fixed during the first 2500 iterations.  The batch size is 16 and each sample in the batch is a 5-frame video clip. The patch size is 128x128. The data augmentation includes random horizontal, vertical, and transposed flipping. 

During the training, we use a two-stage training scheme to  learn the noise-robust feature representation and then make this feature representation help the final  reconstruction. 
In the first stage,  we  use the $ \mathcal{L}oss_{\mathcal{L}2}$ and $\mathcal{L}oss_{CE}$ to train the model for 50K iterations.
The first stage training not only provides the feature generation a good starting point but also helps the compression module converge to a relatively stable status which can effectively filter the noisy and irrelevant information. The second stage only use  $ \mathcal{L}oss_{\mathcal{L}2}$ for the rest iterations.  
With a well-trained compression module, the second stage carefully fine-tunes the quantization step size and the mean value only guided by the $ \mathcal{L}oss_{\mathcal{L}2}$. This helps the model focus more on the model's generation ability for better reconstruction quality.  This simple two-stage training helps the temporal features  be robust to the noise, and then lets these features improve the final quality. 

The main difference of training setting  among different datasets is the number of total iterations.  For DAVIS ~\cite{davis} dataset and CRVD~\cite{crvd} dataset, we train our model for 100K iterations and 200K iterations, respectively. For RainSynLight25~\cite{j4r}, RainSynComplex25~\cite{j4r}, RainSynAll100~\cite{rfmd} and NTU-Rain~\cite{spaccnn}, we train our model for 200K, 200K, 250K and 100K iterations, respectively. For REVIDE~\cite{cgidn}, we train our model for 50K iterations. Besides, we follow the setting in CG-IDN~\cite{cgidn} and use patch size 384x384 in our method for REVIDE. 

\section{More Ablation Studies}
We conduct more ablation studies about the effect of different modules, the weight of the  cross-entropy loss, and the effect of the neural compression-based feature learning on clean features.

\subsection{The Effect of Different Modules}
This paper proposes three key modules: the MV refinement (MVR) , the neural compression-based feature learning (NCFL) incorporated with adaptive quantization, and the feature attention (FA). We study the effect of these modules and report the results in the main paper. 
In this  appendix, we  conduct the experiments by using normal convolutional layers (without auto-encoder structure) to replace  MVR and NCFL, for demonstrating that the improvements of our modules are not from increasing complexity or deeper structure.  MVR is replaced by normal convolutional layers, denoted as M-Conv. NCFL is replaced by normal convolutional layers, denoted as N-Conv. The complexity of M-Conv and N-Conv is slightly higher than that of MVR and NCFL, respectively. 
As Table \ref{ablation_supp} shows, replacing MVR with M-Conv causes a 0.08 dB PSNR drop ($M_b \rightarrow M_c$) and replacing NCFL  with N-Conv  causes a 0.36 dB PSNR drop ($M_d \rightarrow M_e$).  These results verify that the  improvement of performance is not brought by the increasing complexity but comes from our special design.
\begin{table}[t]
	\centering
	\scalebox{0.90}{
		\begin{tabular}{l|cccccccc}
			\Xhline{1.0pt}
			& $M_a$ & $M_b$ & $M_c$ & $M_d$ & $M_e$ & $M_f$ \\
			\hline
			
			\textbf{MVR} &    & \checkmark            &  & \checkmark& \checkmark & \checkmark   \\ \hline
			\textbf{M-Conv} &    &            & \checkmark   &  & &   \\ \hline
			\textbf{NCFL} &    &             & & \checkmark & & \checkmark\\\hline
			\textbf{N-Conv} & & & & &\checkmark & \\\hline
			\textbf{FA} &    &                               &   & & & \checkmark \\\hline
			\textbf{PSNR}    & 29.75 & 29.87& 29.79 &   30.29 &       29.93   & 30.45 \\\hline
			\textbf{GFLOPs} &   534     &    548   &     557  &     613    &       639        &    771 \\\hline  
			
			\Xhline{1.0pt}
	\end{tabular}}
	\caption{The ablation study on different modules. Tested on Set8 ($\sigma = 50$). 
		MVR denotes the MV refinement. NCFL denotes the neural compression-based feature learning. FA denotes the feature attention. M-Conv represents that we use the normal convolutional layers with slightly larger complexity than MVR to replace MVR. N-Conv represents that we use the normal convolutional layers with slightly larger complexity than NCFL to replace NCFL. } \label{ablation_supp}
\end{table}

\subsection{The Weight of   Cross-Entropy Loss}
In the first stage training,
the weight $\lambda$ of the cross-entropy loss $\mathcal{L}oss_{CE}$ controls the intensity  of filtering the noisy information in the temporal features.  
As Table~\ref{weight} shows,  if $\lambda $ is large (e.g., $\lambda $ =  1/256 or 1/512),  some useful information may also be filtered, which degrades performance a bit. 
On the contrary, if $\lambda $ is too small ( e.g., $\lambda $ =  1/4096), the temporal features may  still contain some noisy information and  performance also drops. The optimal $\lambda$ may be related to the noise level, i.e.  different noise levels require different $\lambda$ to achieve the best performance. To avoid too many parameters tuning,  we use 1/2048 as the default value of $\lambda$ for other experiments. In the future, we  will investigate how to adaptively set  $\lambda$ according to the noise level and content characteristic.

\begin{table}[t]
	\centering
	\begin{tabular}{l|ccccc}
		\Xhline{1.0pt}
		$\lambda$ & 1/256 & 1/512 & 1/1024 & 1/2048 & 1/4096  \\
		\hline

		\textbf{PSNR}    &  30.25 & 30.29 &   30.35 &       30.45   & 30.40 \\
		
		\Xhline{1.0pt}
	\end{tabular}
	\caption{The ablation study on the weight $\lambda$  of $\mathcal{L}oss_{Entropy}$.  Tested on Set8 ($\sigma = 50$).  } \label{weight}
\end{table}
\subsection{The Effect of Neural Compression-Based Feature Learning on Clean Features}
Our neural compression-based feature learning is used to filter the noisy and irrelevant information in noisy features, and it will not discard useful information in the clean features.
To verify it,  we use clean frames from DAVIS train-val dataset as input and groundtruth to train the individual  NCFL module with adaptive quantization (NCFL-AdapQ) and the NCFL module without quantization (NCFL-NoQ). We do not  train our complete model because we directly use the current input frame as the input of restoration module. This will enable the restoration module to directly learn the identity mapping when the input is clean, and make the 
temporal feature propagation useless.  
The results show that NCFL-AdapQ reaches PSNR 30.37 dB and NCFL-NoQ reaches PSNR 30.39 dB. The similar results of these two models show that the proposed adaptive quantization has little impact on the clean features.
\begin{table*}[t]
	\centering
	\scalebox{0.70}{
		\begin{tabular}{cccc ccc ccccc}
			\Xhline{1.0pt}
			\textbf{$\sigma$} & \textbf{VNLnet~\cite{vnlnet}} & \textbf{DVDNet~\cite{dvdnet}} & \textbf{FastDVDNet~\cite{fastdvd}} & \textbf{EMVD-L~\cite{emmvd}} & \textbf{EMVD-S~\cite{emmvd}} & \textbf{EDVR~\cite{edvr}} & \textbf{BasicVSR~\cite{basicvsr}} & \textbf{BasicVSR++~\cite{basicvsrpp}} & \textbf{Ours}\\
			\Xhline{1.0pt}
			10    & 35.83/0.9473 & 38.13/0.9679 & 38.71/0.9702 & 38.57/0.9695 & 36.90/0.9512 & 39.23/0.9732 & 39.55/0.9758 & \textcolor{red}{\textbf{39.71}}/\textcolor{blue}{\textbf{/0.9761}}& \textcolor{blue}{\textbf{39.67}}/\textcolor{red}{\textbf{0.9782}} &\\ 
			20    & 34.49/0.9231 & 35.70/0.9470 & 35.77/0.9468 & 35.39/0.9413 & 33.58/0.9023 & 36.33/0.9516 & 36.65/0.9558 & \textcolor{blue}{\textbf{36.75}}/\textcolor{blue}{\textbf{0.9565}} & \textcolor{red}{\textbf{36.78}}\textcolor{red}{\textbf{/0.9596}}&\\ 
			30   & 33.42/0.9086     & 34.08/0.9255 & 34.04/0.9252 & 33.89/0.9210 & 31.94/0.8878 & 34.62/0.9311 & 35.07/0.9389 &
			\textcolor{blue}{\textbf{35.21}}/\textcolor{blue}{\textbf{0.9403}}&\textcolor{red}{\textbf{35.24/0.9435}}& \\ 
			40   & 32.32/0.8974 & 32.86/0.9040 & 32.82/0.9047 & 32.40/0.8941 & 30.66/0.8508 & 33.40/0.9113 & 33.73/0.9207 &
			\textcolor{blue}{\textbf{33.96}}/\textcolor{blue}{\textbf{0.9229}} &\textcolor{red}{\textbf{34.06/0.9267}}& \\ 
			50   & 31.43/0.8761   &  31.85/0.8829 & 31.86/0.8851  & 31.47/0.8747 & 29.88/0.8273 & 32.41/0.8937 & 32.81/0.9046   &\textcolor{blue}{\textbf{32.93}}/\textcolor{blue}{\textbf{0.9056}} & \textcolor{red}{\textbf{33.11/0.9107}}& \\\hline
			GFLOPs  & - & - & 665 & 1106 & 5 & 3089 & 2947 & 3402 & 771 \\ 
			\hline
			\Xhline{1.0pt}
	\end{tabular}}
	\caption{PSNR/SSIM comparison with SOTA video denoising methods on DAVIS testset. The best performance is highlighted in \textcolor{red}{\textbf{red}} (1st best) and \textcolor{blue}{\textbf{blue}} (2nd best). Our method achieves the best SSIM on all noise levels.} \label{davis_result}
\end{table*}
\begin{figure*}
	
	\begin{center}
		\includegraphics[width=1.0\linewidth]{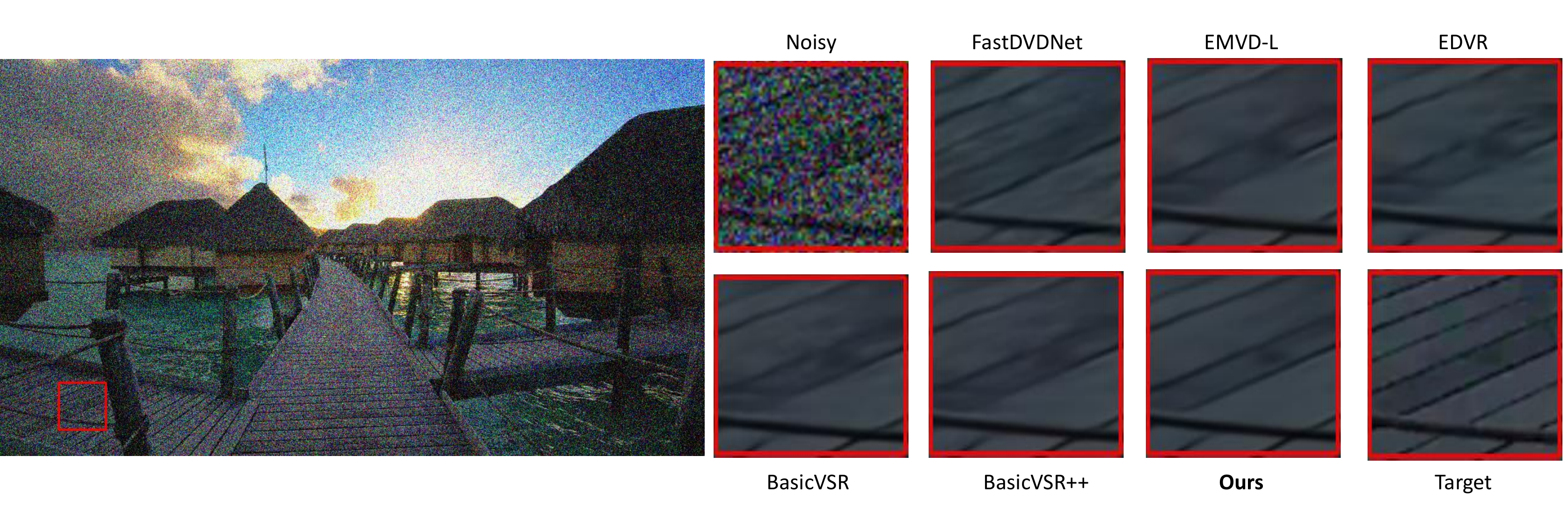}
	\end{center}
	\caption{The denoised results of \textit{hypersmooth}  from Set8 testset   with noise variance 50.  Best viewed in color.  }
	\label{fig0}
\end{figure*}
\begin{figure*}
	
	\begin{center}
		\includegraphics[width=1.0\linewidth]{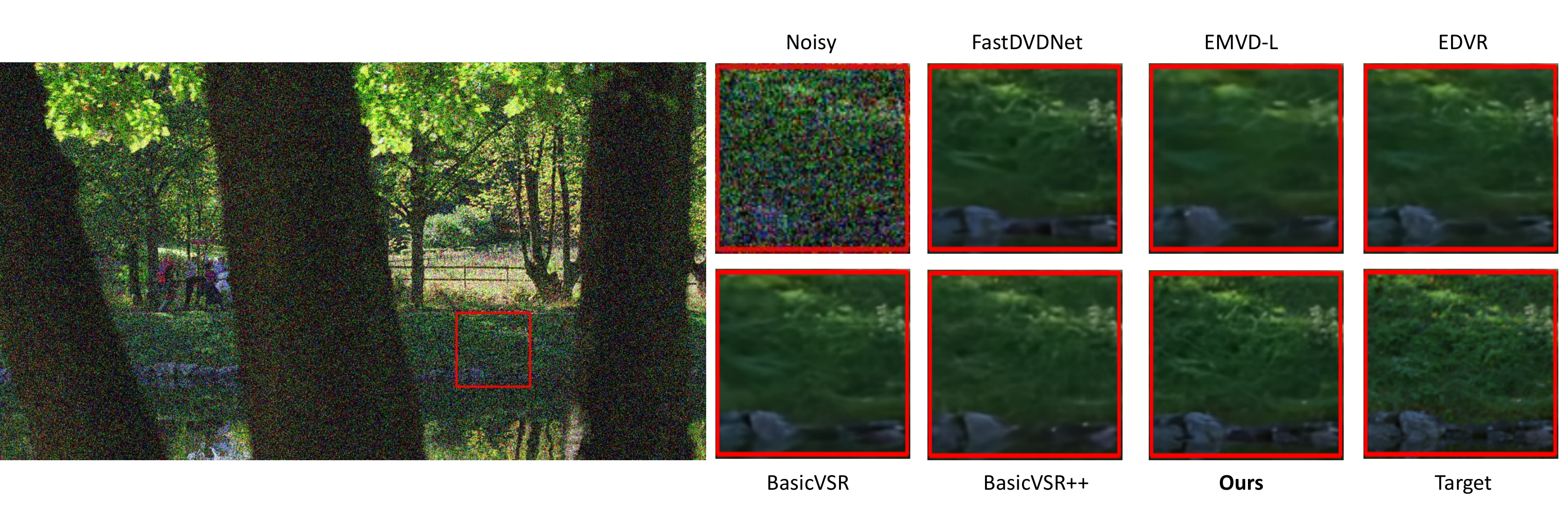}
	\end{center}
	\caption{The denoised results of \textit{park\_joy}  from Set8 test set   with noise variance 50.  Best viewed in color.  }
	\label{fig2}
\end{figure*}

\begin{table*}[t]
	\centering
	\scalebox{0.71}{
		\begin{tabular}{lcc ccc ccccc}
			\Xhline{1.0pt}
			& & \textbf{MS-CSC~\cite{mscsc}} & \textbf{SE~\cite{se}} & \textbf{SpacCNN~\cite{spaccnn}} & \textbf{FastDerain~\cite{fastderain}} & \textbf{J4RNet-P~\cite{j4r} } & \textbf{FCRVD~\cite{fcrvd}} &\textbf{RMFD~\cite{rfmd}} & \textbf{BasicVSR++~\cite{basicvsrpp}} & \textbf{Ours} \\
			\Xhline{1.0pt}
			\multirow{2}*{NTU-Rain} & PSNR &   27.31 & 25.73 & 33.11 & 30.32 & 32.14& 36.05 & 38.92 & \textcolor{blue}{\textbf{39.48}} &\textcolor{red}{\textbf{40.22}} \\ 
			& SSIM &   0.7870   & 0.7614 & 0.9474 & 0.0.9262 & 0.9480 & 0.9676 & 0.9764 & \textcolor{blue}{\textbf{0.9776}} & \textcolor{red}{\textbf{0.9811}} \\ \hline
			
			\multirow{2}*{RainSynLight25} & PSNR &    25.58 & 26.56& 32.78 & 29.42 & 32.96& 35.80 & 36.99 & \textcolor{blue}{\textbf{38.56}} &\textcolor{red}{\textbf{39.17}} \\ 
			& SSIM &   0.8089 & 0.8006 & 0.9239 & 0.8683 & 0.9434 & 0.9622 & 0.9760 & \textcolor{blue}{\textbf{0.9813}} & \textcolor{red}{\textbf{0.9872}} \\ \hline

			\hline
			\Xhline{1.0pt}
	\end{tabular}}
	
	\caption{Comparison with SOTA video deraining methods on NTU-Rain~\cite{spaccnn}  and RainSynLight25~\cite{j4r}. The best performance is highlighted in \textcolor{red}{\textbf{red}} (1st best) and \textcolor{blue}{\textbf{blue}} (2nd best). We train the BasicVSR++~\cite{basicvsrpp} using the same setting as ours. Other baseline results are provided by RMFD~\cite{rfmd} paper.} \label{derain_supp}
\end{table*}

\begin{figure*}[]
	
	\begin{center}
		\includegraphics[width=1.0\linewidth]{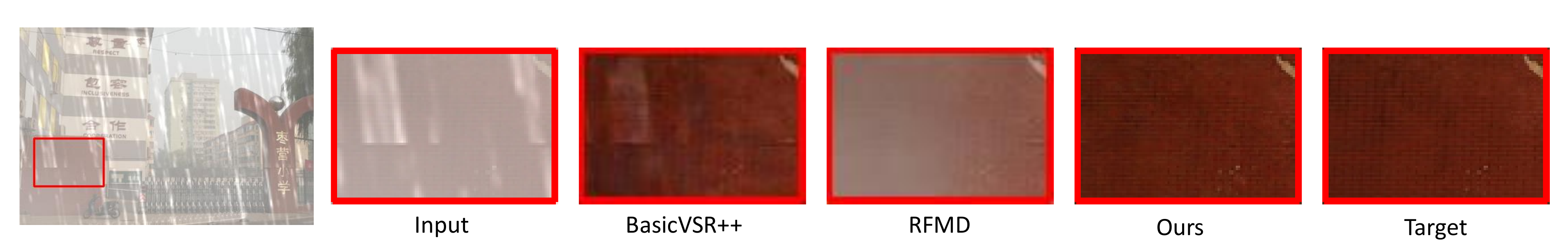}
	\end{center}
	\caption{The  results of \textit{0995} video  from RainSynAll100 testset .  Best viewed in color.  }
	\label{rain_fig1}
\end{figure*}

\begin{figure*}[]
	
	\begin{center}
		\includegraphics[width=1.0\linewidth]{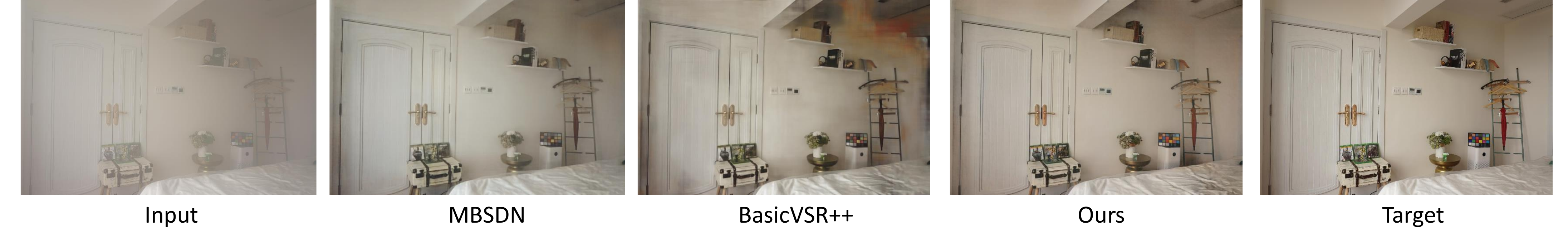}
	\end{center}
	\caption{The  results of \textit{W002} video from REVIDE testset.  Best viewed in color.  }
	\label{haze}
\end{figure*}
	

\section{More Comparisons on Video Denosing}
We compare our method with these baselines:  VNLNet~\cite{vnlnet}, FastDVDNet~\cite{fastdvd}, EMVD~\cite{emmvd}, EDVR~\cite{edvr}, BasicVSR~\cite{basicvsr}, and BasicVSR++~\cite{basicvsrpp}.
EMVD has several network structure configurations with different complexities. More specifically, EMVD mainly contains three modules: the fusion module, the denoising module, and the refinement module. The number of convolutional layers and the  number of the channel of the intermediate features determine the complexity. The default setting of EMVD is (f-2-16, d-2-16, r-2-16). 'f-2-16' means that the fusion module contains 2 convolutional layers and the  number of the channel of the intermediate feature is 16. 'd-2-16' represents the configuration for the denoising module and 'r-2-16' represents the configuration for the refinement module. We denote the default setting as EMVD-small (EMVD-S, f-2-16, d-2-16, r-2-16). In addition, we also compare another  setting, i.e. EMVD-large (EDVR-L, f-4-64, d-6-256, r-4-64). The two settings can be found in the Table 1(b) in EMVD paper~\cite{emmvd}. Since the official code of EMVD is not released, we use the third-party implementation~\cite{emvd_git}.

\begin{table}[t]
	\centering
	\scalebox{0.85}{
		\begin{tabular}{ccccccc}
			\Xhline{1.0pt}
			& $\sigma$=10& $\sigma$=20&  $\sigma$=30&$\sigma$=40 & $\sigma$=50 \\\hline 
			Median filtering (3x3)  & 27.87 & 22.45 & 19.91 &  18.34  & 15.95 \\ 
			Median filtering (5x5) & 26.63 & 21.80 & 19.43 &  17.92  & 15.65 \\ 
			Ours &37.12 & 34.22& 32.57 &   31.39 & 30.45 \\ 
			\Xhline{1.0pt}
	\end{tabular}} 
	\caption{PSNR comparison with median filtering under AWGN, Set8 testset.} 
	\label{mf}
\end{table}

\textbf{Quantitative Comparison.}
In the main paper, we have shown the  results on Set8 testset when compared with previous neural network-based methods. 
In addition, to help us better understand the advantage of neural network-based method over traditional method, we also test median filtering  and summarize the comparison in Table \ref{mf}. From this table, we can see that our method achieves significant quality improvement over median filtering under all noise levels.

In this  appendix, we also provide the PSNR/SSIM results for DAVIS testset~\cite{davis} in Table~\ref{davis_result}. As Table~\ref{davis_result} shows, our method outperforms other methods on high noise levels ($\sigma$ = 20, 30, 40, and 50)  in terms of  PSNR  and outperforms them on all noise levels in terms of SSIM. 
With the increasing of noise level , the  difference of performance  between our method and BasicVSR++  also becomes larger. When $\sigma$ is 10, 20, 30, 40, and 50, the  difference of PSNR between our method and BasicVSR++ is -0.04, 0.03, 0.03, 0.10 and 0.18, respectively. These results demonstrate that the advantage of our neural compression-based feature learning framework is greater when the noise intensity is larger.

\textbf{Qualitative Comparison.}
We show another two visual examples in Fig.~\ref{fig0} and Fig.~\ref{fig2}.
As Fig.~\ref{fig0} shows, other methods cannot restore the bridge deck texture well under this challenging case. The results of FastDVDNet and EMVD suffer from serious distortion and are quite blurry. 
The results of EDVR, BasciVSR, and BasicVSR++ show better visual quality but the bridge deck textures are still not clear. 
By contrast, through leveraging better temporal feature alignment and neural compression-based  feature learning,  our method restores  clearer and more accurate bridge deck textures. 
As Fig.~\ref{fig2} shows, the grass patterns are smoothed in the results of  FastDVDNet, EMVD, EDVR, BasicVSR, and BasicVSR++. The results of our method are clearer and the restored grass patterns are more similar to the target.

\section{More Comparisons on Video Deraining}
We compare our method with prior SOTA video deraining methods, including MS-CSC~\cite{mscsc}, SE~\cite{se},  SpacCNN~\cite{spaccnn}, FastDerain~\cite{fastderain}, J4RNet-P~\cite{j4r}, FCRVD~\cite{fcrvd}, RMFD~\cite{rfmd}, and BasicVSR++~\cite{basicvsrpp}. 
In the main paper, we present the results on RainSynAll100~\cite{rfmd} and RainSynComplex25~\cite{j4r}. In this appendix, we report the results on NTU-Rain~\cite{spaccnn} and RainSynLight25~\cite{j4r}.
As Table~\ref{derain_supp} shows, BasicVSR++ beats RMFD in terms of PSNR and SSIM on both testsets.  Compared with BasicVSR++,  our method even brings PSNR gain of 0.74 dB and SSIM gain of 0.0035 on NTU-Rain, which shows the benefit of the  noise-robust  features. 

We show  visual comparison in Fig.~\ref{rain_fig1}. As  Fig.~\ref{rain_fig1}   shows, BasicVSR++ could not remove the rain streak  well and suffers from severe artifacts. RMFD could remove   rain streak completely but its results suffer from serious color shading. Instead, our method could produce   clearer and  visual pleasing results.

\section{Visual Comparisons on Video Dehazing}
Since the code of CG-IDN~\cite{cgidn} is not  released, we compare our method with BasicVSR++\cite{basicvsrpp} and MBSDN~\cite{mbsdn} in Fig.~\ref{haze}.
As  Fig.~\ref{haze}  shows, the result of BasicVSR++ contains many artifacts and suffers from poor visual quality. The result of MBSDN is cleaner but its  color temperatures are inconsistent with the target. Instead, our method could produce  visual pleasing result with accurate color temperatures.

\end{appendices}

\end{document}